\documentclass{article}
\usepackage{kr}

\usepackage{times}
\usepackage{soul}
\usepackage{url}
\usepackage[hidelinks]{hyperref}
\usepackage[utf8]{inputenc}
\usepackage[small]{caption}
\usepackage{graphicx}
\usepackage{amsmath}
\usepackage{amsthm}
\usepackage{booktabs}
\usepackage{algorithm}
\usepackage{algorithmic}
\usepackage{amssymb}
\usepackage{multirow}
\usepackage{todonotes}

\newtheorem{example}{Example}
\newtheorem{theorem}{Theorem}

\renewcommand{\Re}{\mathbb{R}}
\newcommand{\tnorm}{\theta}
\newcommand{\tconorm}{\kappa}
\newcommand{\fneg}{\nu}

\newcommand{\ontology}{\mathcal{O}}
\newcommand{\ABox}{\mathcal{A}}
\newcommand{\TBox}{\mathcal{T}}

\newcommand{\I}{{\mathcal{I}}}
\newcommand{\ModelName}{\text{F}\mathcal{ALC}\text{ON}}

\newcommand{\Ind}{{\mathbf{I}}}
\newcommand{\Conc}{{\mathbf{C}}}
\newcommand{\Rol}{{\mathbf{R}}}

\newcommand{\ALC}{$\mathcal{ALC}$}

\newtheorem{corollary}{Corollary}
\newtheorem{definition}{Definition}

\title{$\ModelName{}$: Scalable Reasoning over Inconsistent \ALC{} Ontologies}

\author{%
Tilman Hinnerichs\footnote{Equal contributions.}$^{,1,3}$\and
Zhenwei Tang$^{*, 1, 2}$\and
Xi Peng$^{1}$\and
Xiangliang Zhang$^4$ \and
Robert Hoehndorf $^1$ \\
\affiliations
$^1$Computational Bioscience Research Center, Computer, Electrical and Mathematical Sciences \& Engineering Division, King Abdullah University of Science and Technology\and
$^2$University of Toronto \and \\
$^3$Delft University of Technology\and
$^4$University of Notre Dame \\
\emails
t.r.hinnerichs@tudelft.nl\and
josephtang@cs.toronto.edu\and \\
\{xi.peng, robert.hoehndorf\}@kaust.edu.sa\and
xzhang33@nd.edu
}

\begin{document}

\maketitle

\begin{abstract}
Ontologies are one of the richest sources of knowledge. 
Real-world ontologies often contain thousands of axioms and are often
human-made.
Hence, they may contain inconsistency and 
incomplete information which may impair classical reasoners to compute
entailments that are considered as useful.
To overcome these two challenges, we propose $\ModelName{}$, a \underline{F}uzzy \underline{O}ntology \underline{N}eural reasoner to
approximate reasoning over \ALC{} ontologies. 
We provide an approximate technique for the model generation step in classical \ALC{} reasoners. 
Our approximation is not guaranteed to construct exact logical models, but can approximate arbitrary models, which is notably faster for some large ontologies. 
Moreover, by sampling multiple approximate logical models, our technique supports
approximate entailment also over inconsistent ontologies. 
Theoretical results show 
that more models generated lead to closer, i.e., \textit{faithful}
approximation of entailment over \ALC{} entailments.  
Experimental results show that $\ModelName{}$ enables approximate reasoning and reasoning in the presence of inconsistency.  
Our experiments further demonstrate how ontologies can improve knowledge base completion in biomedicine by incorporating knowledge expressed in \ALC{}.
\end{abstract}

\section{Introduction}


\ALC{} ontologies are some of the largest and most expressive
ontologies with thousands of axioms, especially in the bio-medical
domain \cite{smith2007obo}.
However, they are also human-made, leading to both possible
inconsistencies and incomplete information.

Classical logical reasoners are limited in this case: 
First, they do not compute \textit{useful} inferences in the presence
of contradictions as all statements are entailed as they logically
should. 
Even if the contradiction is small and in a remote branch of the ontology, it makes the rest of the ontology unusable until the inconsistency is resolved.
Correcting contradictions is either infeasible for large KBs or done such that potentially valuable information is deleted \cite{troquard2018repairing}.
This is not \textit{useful} not in practice.
Second, as reasoning in \ALC{} is \textsc{EXPTIME}-complete, reasoning
becomes computationally infeasible for large and complex ontologies.

Recently, neural entailment methods have been developed to deal with
larger ontologies and missing information \cite{xiong2023geometric}.
These methods replace symbols with embeddings and map reasoning to algebraic operations over these embeddings.
For ontologies specifically, concepts are interpreted as regions in
form of geometric shapes, e.g. balls \cite{kulmanov2019embeddings} and
boxes \cite{xiong22_iswc}.
While they are capable to handle inconsistencies, they either cope with only some aspects of semantic entailment such as logical query answering for specialized queries \cite{hamilton2018embedding,ren2020query2box} or limit the scope to simpler logics \cite{kulmanov2019embeddings,xiong22_iswc,jackermeier2023box}.

Logic Tensor Networks (LTNs) \cite{badreddine2022logic} combine real logic and neural networks to reason over vector spaces yielding the expressivity of first-order logic (FOL). 
However, LTNs scale poorly if many predicates of arity larger than $1$ and constants are present \cite{bianchi2019capabilities}, which is the case in many real world ontologies.
LTNs are hence not suitable for such reasoning tasks.

We propose $\ModelName{}$: \underline{F}uzzy
$\underline{\mathcal{ALC}}$ \underline{O}ntology \underline{N}eural
Reasoner for approximate and scalable reasoning over \ALC{}
ontologies, using two ideas:
First, we generalize the notion of geometric shapes and approximate arbitrary membership functions of concepts.
This allows for arbitrary interpretations of concepts.
We further use fuzzy logic operators to express complex concept descriptions and approximate arbitrary, logical models for full \ALC{} ontologies.
Second, we generate multiple approximate models to enable \textit{approximate entailment} and describe its semantics. 

$\ModelName{}$ reasons under the open-world assumption, i.e. does not assume the truth value of a statement when it is not derived yet. 
This is useful for describing knowledge in a way that is extensible and hence core to ontologies.

$\ModelName{}$ also gives rise to other forms of approximate entailment.
Consider the example $Father \sqsubseteq Parent$ in the absence of $Father \sqsubseteq Male$ where we aim to entail $Father \sqsubseteq Parent \sqcap Male$.
It is not classically entailed, even if all named instances of
$Father$ are also instances of $Male$ due to the open world assumption
(supported in $\ModelName{}$). 
If all instances of $Father$ are also instances of $Male$ only considering named individuals then $Father \sqsubseteq Male$ would be true in all models generated by $\ModelName{}$.

We summarize the main contributions of this work as:
\begin{itemize}
\item We propose and implement the first neural approach for computing approximate semantic entailment with the full expressivity of \ALC{} ontologies; 
\item We obtain experimental results showing 
computing approximate semantic entailments with $\ModelName{}$ enables approximate reasoning in the presence of inconsistencies as well as discovery of facts in large \ALC{} ontologies;
\item We prove that $\ModelName{}$ can represent any logical model (representation completeness) and that the objective function generates logical models if it reaches an arbitrarily small threshold (faithfulness). We hence prove that $\ModelName{}$ approximates semantic entailment.
\end{itemize}

\section{Preliminaries}
\subsection{Description Logic \ALC}
Description logics (DLs) are fragments of first-order logic. A DL
ontology consists of a set of individual symbols $\Ind=\{a,b,...\}$, a
set of concept symbols $\Conc=\{C,D,...\}$, and a set $\Rol=\{R,...\}$
defining relations between individuals, summarized in signature
$\Sigma=(\Ind, \Conc, \Rol)$.

The DL \ALC{} \cite{baader2003description} constructs \textit{concept
  descriptions} extending the combinators intersection $C \sqcap D$
and existential quantifier $\exists R.C$, also present in the simpler
DL $EL$, by adding negation $\neg C$, union $C \sqcup D$, and
universal quantifiers $\forall R.C$. Facts in \ALC{} are divided in
TBox and ABox axioms.  A TBox consists of axioms of the form
$C \sqsubseteq D$, i.e. all members of $C$ must also be in $D$, and an
ABox axiom the form $C(a)$ or $R(a,b)$, i.e. $a$ is member of $C$, or
$R$ relates $a,b$.

\begin{example}
    \label{ex:father_ont}
    We model fatherhood with ontology $\ontology = \ABox \cup \TBox$
    with ABox $\ABox=\{Father(dave)\}$ and TBox
    $\TBox=\{ Father \sqsubseteq \exists hasChild.Person\}$.  Here,
    $\Ind=\{dave\}, \Conc = \{Father, Person\}$ and
    $\Rol = \{hasChild\}$ form signature $\Sigma$.
\end{example}

The axioms $Person \sqsubseteq Father \sqcup Mother$ and $Mother
\sqsubseteq \neg Father $ can not be expressed in $EL$ but in \ALC{}.

An interpretation $\I$ maps symbols in $\Sigma$ to (sets of) elements
of a non-empty universe $\Delta^\I$ using interpretation function
$\cdot^\I$ (for an intuition, see Example
\ref{ex:father_model}). $\cdot^\I$ assigns all individual names
$a\in\Ind$ to elements of $\Delta^I$ such that $a^\I \in \Delta^\I$,
all concepts $C\in\Conc$ to subsets $C^\I \subseteq \Delta^\I$, and
all relations $R\in\Rol$ to relations of subsets
$R^\I \subseteq \Delta^\I \times \Delta^\I$.

The interpretation function is extended to \textit{concept descriptions}, i.e. combinations of relations and concepts, as:
\begin{equation}
\label{equ:concept_descriptions}
\small
\begin{split}
    &(C \sqcap D)^\I := C^\I \cap D^\I, \ (C \sqcup D)^\I := C^\I \cup D^\I, \\
    &(\forall R.C)^\I := \{ d \in \Delta^\I | \forall e \in
\Delta^\I: (d,e) \in R^\I \mbox{ implies } e \in C^\I\}, \\
    &(\exists R.C)^\I:= \{ d \in \Delta^\I | \exists e \in
\Delta^\I: (d,e) \in R^\I \mbox{ and } e \in C^\I\}, \\
    &(\neg C)^\I := \Delta^\I - C^\I.
\end{split}
\end{equation}

Interpretations do not have to satisfy the axioms. Satisfying interpretations are called \textit{logical models}.
\begin{example}
    \label{ex:father_model}
    A satisfying interpretation, i.e. logical model, $\I$ of $\ontology$ is defined by universe $\Delta^\I = \{d,e\}$ and assignments $dave^\I=d$, $Father^\I = \{d\}$ and $hasChild=\{(d,e)\}$. Further, $(\exists hasChild.Person)^\I=\{d\}$.
\end{example}

Note that interpretations of concepts and concept descriptions can be described by a membership function $m_C(e)\mapsto \{0,1\}$ for $e\in \Delta^\I$, e.g. with $m_{Father}(d)=1$.
Finally, \ALC{} ontologies (and all subsets) have the \textit{finite model property}, i.e., all non-theorems can be falsified by a finite model.

\section{Methods}
Reasoning over \ALC{} ontologies can be done by constructing a single
satisfying interpretation, reducing entailment to (non-)satisfiability
\cite{tsarkov2006fact++}.  We refer to \cite{baader2003description}
for more information and definitions.
As this reasoning is \textsc{EXPTIME}-complete, it can  infeasible for large and complex ontologies, or in case of an inconsistent KB, not always \textit{useful} as all statements are entailed. 
For inconsistencies, we aim to derive facts that \textit{should be entailed} while preserving all expert domain knowledge and not resolving the contradiction.  

Similarly to classical reasoners, we first describe how to approximate
a logical model for an \ALC{} ontology.
To do so we first construct the underlying universe, define an
interpretation and finally optimize the interpretation for
satisfaction of the ontology.

In contrast to most classical reasoners, we do not reduce entailment to satisfiability. 
Instead we reason over multiple sampled approximate models in order to perform approximate entailment.


\subsection{Generating model structures}
\label{generating_models}

We generate interpretations using two ideas: First, and similar to
other neural entailment methods,
we embed symbols of signature $\Sigma$ into $\Re^n$ using an embedding
function $f_e$.  Second, we learn the satisfying interpretation of a
concept by approximating its membership function using fuzzy sets as a
continuous relaxation with $m_C:\Delta^\I\mapsto [0,1]$ for concept
$C$ and interpretation $\I$.  Concepts hereby refer to both concept
symbols and concept descriptions.

\subsubsection{Constructing the Universe}

To construct an interpretation we first define the universe.
In order to consider arbitrarily large universes, we extend the set of individuals by our embedding space: 
Given the signature $\Sigma = (\Conc, \Rol, \Ind)$ of ontology
$\ontology$, we add to $\Ind$ a set of individual symbols
$\Ind_{\Re^n}$ where $\Ind_{\Re^n}$ contains one new individual symbol
for every member of the set $\Re^n$.
We set the interpretation of $x^\I=x$ for $x\in\Ind_{\Re^n} $ and $a^\I=f_e(a)$ for $a\in\Ind - \Ind_{\Re^n} $.
Our interpretation is now defined over the universe $\Delta$ consisting of all individual names plus the set of sampled individuals, leading to $\Delta^\I = \Re^n$ for all $\I$.

We now have the challenge that, for logical concept interpretations,
we must provide a degree of membership for each element of an
uncountable set of individuals. 
Inspired by previous work \cite{van2022analyzing}, we address this problem by sampling a (finite) set of individual symbols from $\Ind_{\Re^n}$, in addition to all the individual symbols from $\Ind$, and applying the approximated membership function $m_C(\cdot)$. 

For inference, i.e., the prediction, the degree of membership of any element of $\Ind_{\Re^n}$ 
in any concept description $C$ can be computed directly through its membership function $m_C$.
Further, it suffices to sample a finite number of individuals to find logical models due to the finite model property of \ALC. 
The sampling strategy can be either uniform or constrained in various
ways (e.g., by sampling in the proximity of embeddings of named
individuals).


\subsubsection{Interpreting Concept Symbols}

Concepts are a core notion in constructing a logical model of
an ontology. 
For a concept $C$, we approximate
the notion of membership from $m_C:\Delta \mapsto \{0,1\}$ by using
fuzzy sets with membership function $m_C^:\Delta \mapsto [0,1]$.

With the embedding function $f_e:\Sigma \rightarrow \Re^n$, we interpret a concept symbol $C\in\Conc$ as an MLP modeling its membership with
\begin{equation}
\small
 m_C(x) = \sigma(MLP(f_e(C), f_e(x)))
  \label{eqn:fsconcept}
\end{equation}
where $f_e(C)$ is the embedding of $C$, $f_e(x)$ the
embedding of the individual $x$, $\sigma$ the sigmoid
function, and a multilayer perceptron (MLP). Note that all membership functions share the same $MLP$.

As we aim to find not just interpretations but satisfying ones, i.e. logical models, of an ontology $\ontology$ we denote this function with $f_{mod}(C)$, fully defined by its membership function $m_C$. Note that without optimization $m_C$ only returns a interpretation, not a model.




\subsubsection{Relation Symbols}
To tackle arbitrary \ALC{} axioms, we also have to define an interpretation for relation symbols $R\in\Rol$, which describe relations between pairs of elements.
The interpretation of a relation symbol $R$ hence relates two individuals, assigning a degree of membership to a set of tuples of elements written as the pair $(m, \Delta \times \Delta)$ which is here defined as:
\begin{equation}
\small
  m_R((x,y)) = \sigma(MLP(f_e(x) + f_e(R), f_e(y)))
    \label{eqn:fsrelation}
\end{equation}
based on TransE~\cite{bordes2013translating}.

\subsubsection{Concept Descriptions}
We can now combine the interpretations of concept and relation symbols to complex concept descriptions. 
Crucially, we have to define how to combine two fuzzy membership functions over the operators defined in \ALC.

We use $t$-norm $\theta$ \cite{van2022analyzing} (common choice is the product) to describe fuzzy intersection, $\kappa$ for the corresponding $t$-conorm to describe union, and fuzzy negation $\fneg$. We assume $\theta$ to be continuous and differentiable, where we list additional formal properties and requirements in the Appendix.

Interpretations of a concept description $C$ are defined recursively. 
For the operators intersection $\sqcap$, union $\sqcup$, and negation $\neg$, we define the membership function $m$ as
\begin{equation}
\small
  m_{C_1 \sqcap C_2}(x) = \tnorm(m_{C_1}(x), m_{C_2}(x))
  \label{eqn:intersection}
\end{equation}
\begin{equation}
\small
  m_{C_1 \sqcup C_2}(x) = \tconorm(m_{C_1}(x), m_{C_2}(x))
  \label{eqn:union}
\end{equation}
\begin{equation}
\small
m_{\neg C}(x) = \fneg (m_C(x))
  \label{eqn:negation}
\end{equation}
for concept descriptions $C_1, C_2, C$.


For the fuzzy membership of $x$ in $\exists R.D$ (see Eq. \ref{equ:concept_descriptions}),
we iterate through the set of individuals in $\Delta$ to find the
maximum of memberships in $D^\I$ that stand in relation $R$ to $x$
(i.e., membership of $(x,y)$ in $R^\I$) using $\tnorm$:
\begin{equation}
  \small
  m_{\exists R.D}(x) = \max_{y \in \Delta } \tnorm ( m_{D}(y),
  m_R((x,y)))
    \label{eqn:existential}
\end{equation}
For the universal quantifier, we follow a similar approach and assign
the degree of membership of $x$ in $(\forall R.D)^\I$ (see
Eq. \ref{equ:concept_descriptions}) as:
\begin{equation}
  \small
  m_{\forall R.D}(x) = \min_{y \in \Delta }  \tconorm
  (\fneg(m_R((x,y))) , m_D(y))
    \label{eqn:universal}
\end{equation}
In these formulations, the choice of $t$-norm, $t$-conorm, and
fuzzy negation are hyperparameters of our method. 

\subsubsection{Optimization}
We now defined how to find an interpretation of ontology $\ontology$,
that we now want to transform to a satisfying one. An interpretation
is a logical model if it satisfies all axioms $C \sqsubseteq D$
(satisfied if $C^\I \subseteq D^\I$) of TBox $\TBox$ and all axioms
$C(e)$ and $R(e_1, e_2)$ of ABox $\ABox$, i.e. our ground-facts. Note
that we only find a model if the loss is zero,
which we elaborate in Section \ref{sec:theoretical_results}.

We first approach the problem of finding a model of TBox axioms $C \sqsubseteq D$, with concept descriptions $C,D$.  
First, we normalize the TBox and rewrite all axioms of the type $C \sqsubseteq D$ as $C \sqcap \neg D \sqsubseteq \bot$ as $C^\I\subseteq D^\I$ iff $(C \sqcap \neg D)^\I\subseteq\emptyset$.
We express this goal with membership functions by minimizing the degree of membership in $m_{(C \sqcap \neg D)}$ for any possible individual.

With this formulation, given a set of individual symbols $E$ sampled from $\Ind \cup \Ind_{\Re^n}$ and given the TBox $\mathcal{T}$, the degree of membership of entities in each of the disjuncts can be minimized through this loss:
\begin{equation}
\small
    \mathcal{L}_{\mathcal{T}} = 
    \frac{1}{| E |}
    \frac{1}{|\mathcal{T}|} 
    \sum_{C \sqsubseteq D \in \mathcal{T}}\sum_{e \in E} m_{(C \sqcap \neg D)}(e)
\end{equation}
This loss ensures that TBox axioms are satisfied in the interpretation
generated, i.e. we generate a logical model. We normalize each loss to allow for a weighted loss formulation in Eq. \ref{eqn:loss}.

A second loss is defined for ABox axioms of the type $C(e)$ (concept assertion). It aims to ensure that all given facts of type $C(e)$ are satisfied and will maximize the degree of membership of $e$ in $C^\I$:
\begin{equation}
\small
    \mathcal{L}_{\mathcal{A}_1} = 
     \frac{1}{|\mathcal{A}_1|} 
    \sum_{C(e) \in \mathcal{A}_1} (1-(m_C(e)))
\end{equation}
A third loss for ABox axioms of type $R(e_1, e_2)$ (role assertions) maximizes membership of pairs of individuals $(e_1, e_2)$:
\begin{equation}
\small
    \mathcal{L}_{\mathcal{A}_2} = 
    \frac{1}{|\mathcal{A}_2|} 
    \sum_{R(e_1, e_2) \in \mathcal{A}_2} (1- m_R((e_1,e_2)))
\end{equation}
The final loss function is:
\begin{equation}
\small
  \mathcal{L} = \alpha \mathcal{L}_{\mathcal{T}} + \beta
  \mathcal{L}_{\mathcal{A}_1} + (1 - \alpha - \beta)
  \mathcal{L}_{\mathcal{A}_2}
\label{eqn:loss}
\end{equation}
with $\alpha, \beta\in [0,1]$ and $\alpha + \beta < 1$.

The algorithm randomly initializes the two learnable components $f_e$ and $MLP$ and minimizes loss $\mathcal{L}$ through gradient descent. 
The choice of $t$-norm and $t$-conorm as well as fuzzy negation are parameters of the algorithm, and any differentiable $t$-norm and $t$-conorm \cite{van2022analyzing} can be used.
Furthermore, the number of individuals to sample from the embedding space ($\Ind_{\Re^n}$), and the sampling strategy, are parameters of the algorithm.

To summarize, we are able to approximate one model of the input ontology $\ontology$, by describing a membership function for arbitrary \ALC{} axioms, and then try to correct it into a satisfying logical model by optimization over our loss function.

\subsection{Faithful semantic entailment}
\label{multiple_model_entailment}
Given the ability to represent and approximate single logical models of ontologies, we now outline how to reason over multiple of such models. 

Classical \textit{semantic entailment} $\models$ of a TBox axiom $C \sqsubseteq D$ with concept descriptions $C,D$ is defined as a relation between classes of models.
Specifically, all logical models (satisfying interpretations) $Mod(\TBox)$ of TBox $\TBox$ have to be models of $C\sqsubseteq D$, denoted by $\TBox \models C\sqsubseteq D$.
In general, $Mod(\TBox)$ will be a class, and thus will be impossible to enumerate. 

However, as \ALC{} has the finite model property, any non-theorem is falsified by some finite model bounded in size of the ontology.
Hence, we can sample from $Mod(\TBox)$ restarting our algorithm to construct multiple models to disprove more axioms until a model is disproven. 
Thus, an approximate form of semantic entailment arises. 

\subsubsection{Approximate Reasoning}
We re-formulate common reasoning tasks \cite{bobillo2016fuzzy} over fuzzy ontologies, using membership functions.
We further give intuition that $\ModelName{}$ converges to entailment the more models we generate for the tasks (but not limited to) concept satisfiability, instantiation reasoning, subsumption and consistency.
We show this relation in Section \ref{sec:theoretical_results}.

Satisfiability of a concept description $C$ using the degree of membership is defined as:
\begin{definition}[Concept satisfiability]
    A concept $C$ is called satisfiable w.r.t TBox $\TBox$ if
    there is a model $\I$ of $\TBox$ and an $a \in \Delta^\I$ such that
    $m_C(a) = 1$. 
\end{definition}

Degree of satisfiability relaxes this formulation to fuzzy satisfaction and is equivalent in the case $\alpha=1$.
\begin{definition}[Approximate degree of satisfiability]
    Let $Mod_k(\TBox)$ be the set of $k$ generated models of
    $\TBox$ by $\ModelName$.  $C$ is satisfiable to degree $\alpha$ w.r.t TBox
    $\TBox$ and $Mod_k(\TBox)$ 
    if $\alpha = \max_{\I \in Mod(\TBox)} \max_{a \in \Delta^\I}
    m_C(a)$.
\end{definition}

We can use the degree of satisfiability to define truth values of subsumptions between concept descriptions. 
\begin{definition}[Approximate truth value of subsumption]
  Let $C$ and $D$ be concept descriptions. The truth value $\alpha$ of
  the subsumption statement $C \sqsubseteq D$ w.r.t TBox $\TBox$ and $Mod_k(\TBox)$ is
  the degree of satisfiability of $\neg C \sqcup D$ w.r.t $\TBox$ and $Mod_k(\TBox)$,
\end{definition}


A similar approach can be used to formulate the instantiation reasoning task, i.e., the task of finding all (named) individuals that are the instance of a class description.
\begin{definition}[Approximate concept instantiation]
  We define $\TBox \cup \ABox \models_k^\alpha C(a)$ as
  $\min_{\I \in Mod_k(\TBox \cup \ABox)} m_C(a)\geq\alpha$, i.e., the
  minimum degree of membership of $a$ in the interpretation of $C$
  across all $k$ models.
\end{definition}
Approximate entailment of relation assertions can be formalized similarly.

We can eventually test for consistency of an ABox by defining the approximate degree of consistency.
\begin{definition}[Approximate degree of consistency]
  ABox $\ABox$ is consistent with respect to TBox $\TBox$ to degree
  $\alpha$ iff
  $\alpha = \max_{\I \in Mod_k(\TBox)} \min(\{m_C(a) | C(a)\in
    \ABox\} \cup
  \{m_R((a,b))| R(a,b) \in \ABox\})$.
\end{definition}

This leads to our concluding definition of approximate entailment:
\begin{definition}[$(M, \alpha)$-approximate entailment]
  Let $M = Mod_k(\TBox \cup \ABox)$ be $k$ approximate models
  generated by $\ModelName{}$ from $\TBox \cup \ABox$.  If query $\phi$ is
  true to at least degree $\alpha$ in all $\I \in M$, $\phi$ is
  $(M, \alpha)$-approximately entailed by $\TBox \cup \ABox$.
\end{definition}



\section{Theoretical results}
\label{sec:theoretical_results}

We prove that $\ModelName{}$ approximates semantic entailment. 
For this purpose, we first prove \textit{faithfulness}, i.e., that our objective function actually approximates finding a classical model.
Based on this result, we define approximate models and show that any (finite)
classical model can be ``represented'' by $\ModelName{}$ (representation completeness).
This allows us to define our notion of approximate semantic entailment.
We include the proofs in the Supplementary Materials.

\setcounter{theorem}{0}
\begin{theorem}[Faithfulness]
  \label{theorem:loss}
  Let $\TBox$ be a TBox and $\ABox$ an ABox in \ALC{} over signature
  $\Sigma = (\mathbf{C}, \mathbf{R}, \mathbf{I} = \mathbf{I_n} \cup
  \mathbf{I_{\Re^n}})$. If $\mathcal{L}=0$ and the $t$-norm $\tnorm$
  satisfies $\tnorm(x,y)=0$ if and only if $x=0$ or $y=0$, then the
  interpretation $\I$ with $C^\I = f_{mod}(C)$ for all
  $C \in \Conc$, $R^\I = f_{mod}(R))$ for all
  $R \in \Rol$, and $a^\I = a$ for all $a \in \Ind$ is a
  classical model of $\TBox$ and $\ABox$ with domain/universe $\Delta = \Re^n$.
\end{theorem}

The theorem shows that our loss is designed, in the limit, to
construct a logical model. 
If the loss is not $0$, we can define the structure identified as an \textit{approximate model}:
\begin{definition}[Approximate model]
  If $\mathcal{L}>0$, the interpretation $\I$ with
  $C^\I = f_\I(f_e(C))$ for all $C \in \mathbf{C}$,
  $R^\I = f_\I(f_e(C))$ for all $R \in \mathbf(R)$, and $a^\I = a$ for
  all $a \in \mathbf{I}$ is an approximate model of $\TBox$ and
  $\ABox$ with domain $\Delta = \Re^n$.
\end{definition}

We further show that every logical model can be represented by our
algorithm, i.e., that we are able to approximate arbitrarily closely a
loss of $0$ for any finite logical model (representation completeness).
\ALC{} has the ``finite model property'', i.e., every statement that may be false in a model is already false in a finite model. 
Therefore we focus only on finite models (all other models are elementary extensions of these).
\begin{theorem}[Representation completeness]
  \label{thm:complete}
  Let $\TBox$ be a TBox and $\ABox$ an ABox in \ALC{} over signature
  $\Sigma = (\mathbf{C}, \mathbf{R}, \mathbf{I} = \mathbf{I_n} \cup
  \mathbf{I_{\Re^n}})$. If $\I$ is a finite model of
  $\TBox \cup \ABox$ then there exists an $f_e$ and $f_\I$ such that
  $\mathcal{L}\leq \epsilon$ for any $\epsilon>0$.
\end{theorem}

Crucially, Theorem \ref{thm:complete} does not show how to find the
functions $f_e$ and $MLP$, but only shows their existence, i.e., that
every (finite) interpretation is representable using $\ModelName{}$.

There is a clear relation between $(M,\alpha)$-approximate entailment
and semantic entailment: 
\setcounter{theorem}{0}
\begin{corollary}[Relation to entailment]
  If $\phi$ is semantically entailed by $\TBox \cup \ABox$, then
  $\ModelName{}$ can generate approximate models where $\phi$ is true
  to degree $\alpha$ for any $\alpha \leq 1$ (by Theorem
  \ref{theorem:loss}).
  If $\phi$ is {\em not} entailed by $\TBox \cup \ABox$,
  $\ModelName{}$ can generate a model where $\phi$ is false to degree
  $\alpha$ ($\alpha \geq 0$) (by Theorem \ref{thm:complete}).
\end{corollary}

We thus say that $\ModelName{}$ faithfully approximates entailment. 
Approximate entailment is equivalent to classic semantic entailment if
$M=Mod(\TBox \cup \ABox)$ and $\alpha = 1$.

\section{Experiments}

We conduct extensive experiments to answer the following research
questions:
\textbf{RQ1:} Can $\ModelName{}$ perform (approximate) semantic
entailment?
\textbf{RQ2:} Is $\ModelName{}$ robust to inconsistency?
\textbf{RQ3:} Can $\ModelName{}$ improve knowledge base completion in
the biomedical domain by incorporating background knowledge from \ALC{}
ontologies?

\subsection{Experimental settings}
\subsubsection{Datasets}
We use four ontologies for evaluation with special properties for different use cases. 
(1) \textbf{Family ontology}, defined in Eqn. 1 in Supplementary Materials, to test the open-world reasoning.
(2) \textbf{Pizza
  Ontology}\footnote{https://protege.stanford.edu/ontologies/pizza/pizza.owl}
for para-consistent reasoning by, e.g., generating an individual that
is both a Meaty Pizza and a Vegetarian Pizza (which are disjoint in the Pizza Ontology).
(3) Human Phenotype Ontology (\textbf{HPO}) \cite{kohler2021human} for
scientific fact discover over large human-made \ALC{}-ontologies. We
include gene-to-phenotype
annotations\footnote{https://hpo.jax.org/app/download/annotation} in
the form of $\exists anno.Phenotype(gene)$ to the ABox, and we use
gene--gene interactions (GGI) in
BioGRID\footnote{https://downloads.thebiogrid.org/BioGRID}\cite{oughtred2021biogrid}
in the form of $interacts(gene_1, gene_2)$ in the ABox.
(4) \textbf{Yeast}
dataset\footnote{https://bio2vec.cbrc.kaust.edu.sa/data/elembeddings/el-embeddings-data.zip}
for predicting protein--protein interactions (PPI) to compare with
other ontology embedding methods
\cite{kulmanov2019embeddings,xiong22_iswc}.

The first three ontologies are used to evaluate the semantic
entailment performance of $\ModelName{}$. We evaluate Family with
specially prepared axioms as
Eqn.~\ref{eq:family_entailed_1}--\ref{eq:family_undecidable}. We apply
the \ALC{}-reasoner HermiT \cite{shearer2008hermit} on Pizza and apply
$EL$-reasoner ELK \cite{kazakov2014incredibleELK} on HPO to generate
axioms that should be entailed, i.e., \textit{entailments}. Since
automated reasoners do not directly output disproved axioms, we regard
the axioms that are neither included in the original ontology nor
entailed as unprovable axioms. Statistics of datasets can be found in
Table 1 in Supplementary Materials.

\subsubsection{Implementation Details}

We use ranking metrics to evaluate $\ModelName{}$ for GGI on HPO and for PPI on Yeast. 
For GGI, we evaluate on randomly selected testing triples under the filtered setting \cite{bordes2013translating}.
For PPI, we use both the raw (R) and filtered (F) settings \cite{xiong22_iswc} and report on the same train--test split as \cite{kulmanov2019embeddings,xiong22_iswc}.

For GGI and PPI predictions, we apply uniform negative sampling for
each ABox axiom and we use BPR loss \cite{rendle2012bpr} for ABox
optimization on HPO and Yeast. The initial learning rate of the Adam
\cite{kingma2014adam} optimizer, the embedding dimension $d$, and the
number of negative samples for each positive sample in ABox, are tuned
by grid searching within \{$1e^{-2}$, $1e^{-3}$, $1e^{-4}$\}, \{32,
64, 128\}, and \{4, 8, 16\}, respectively.  We set the number of
models $k$ and the choice of $t$-norms as hyperparameters; we use the product
$t$-norm in all experiments.

We sample anonymous individuals before each optimization step;
consequently, we can potentially cover large parts of the embedding
space as the number of optimization steps grows, even with a small
number of anonymous individuals being sampled each time. In our
experiments, we choose to sample two anonymous individuals based on
named individuals with $0$-\textit{mean} $0.1$-\textit{std} Gaussian
noise, and another two anonymous individuals which are sampled using
Xavier initialization \cite{glorot2010understanding}. The tuned
hyperparameters are listed in Table 2 in Supplementary Materials.

\subsection{Experimental results}

\subsubsection{Approximate semantic entailment (RQ1)}
\begin{figure}[!t]
	\centering  
	\includegraphics[width=0.48\textwidth]{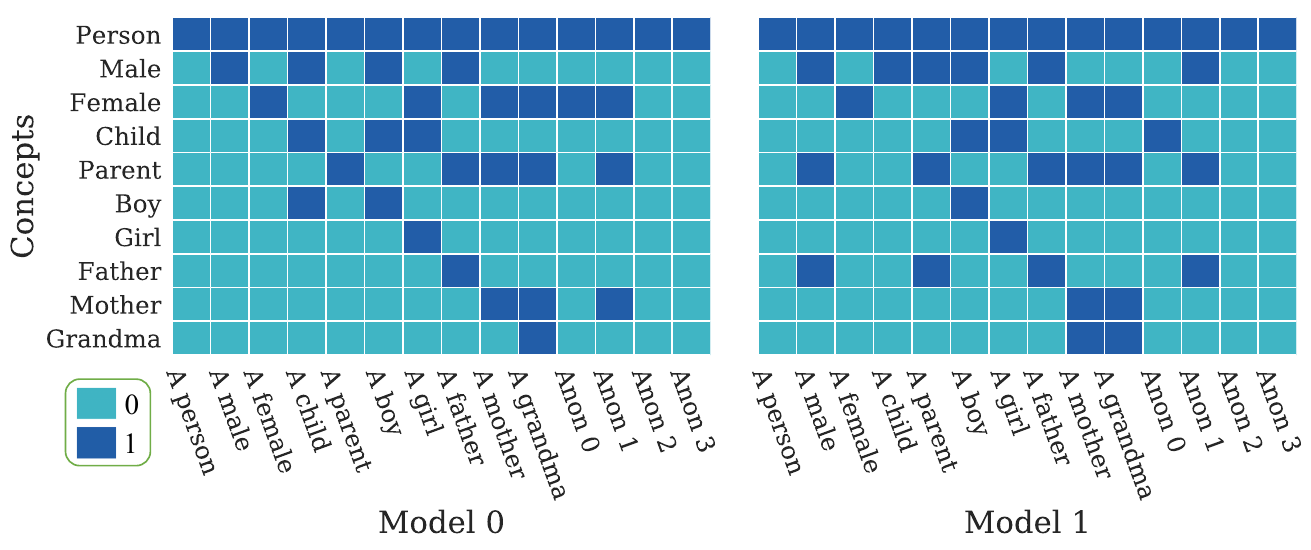}
        \caption{Learned degree of memberships of individuals in
          concepts in the \textbf{Family}. Dark and light
          grids denote the learned degrees of membership are (very
          close to and numerically indistinguishable from) 1 and 0,
          respectively. Anon $i$ denotes the randomly sampled
          anonymous individuals.}
  \label{fig:family}
\end{figure}

We use representative axioms for evaluations on \textbf{Family}:
\begin{equation}
\small
    \text{Female} \sqcap \text{Child} \sqsubseteq \text{Girl},
\label{eq:family_entailed_1}
\end{equation}
\begin{equation}
\small
    (\exists \text{hasChild.Person} \sqcap \text{Female}) \sqsubseteq \text{Mother}
\label{eq:family_entailed_2}
\end{equation}
\begin{equation}
\small
    \text{Person} \sqsubseteq \ \text{Parent},
\label{eq:family_disproved}
\end{equation}
\begin{equation}
\small
    \text{Mother} \sqsubseteq \ \text{Grandma},
\label{eq:family_undecidable}
\end{equation}
where Eqn.~\ref{eq:family_entailed_1} is an axiom that should be
entailed by $\ModelName{}$; we expect the membership within the fuzzy
sets generated from Eqn.~\ref{eq:family_entailed_1} to be close to $0$
in all models.  More intuitively, the intersection of the fuzzy sets
(rows in Fig.~\ref{fig:family}) of \textit{Female} and \textit{Child}
are a subset of the fuzzy set of \textit{Girl} in both
models. Furthermore, $\ModelName{}$ is also capable of entailing more
complex axioms like Eqn.~\ref{eq:family_entailed_2}.
Eqn.~\ref{eq:family_disproved} should be disproved because $Parent$
and $Child$ are disjoint, both are subclasses of $Person$, and there
is an instance of $Child$ in the Family ontology. All models (i.e.,
both Model 0 and Model 1 in Fig.~\ref{fig:family}) make the statement
false.  Thus Eqn.~\ref{eq:family_disproved} is disproved by
$\ModelName{}$.  Eqn.~\ref{eq:family_undecidable} is false in model
$0$ while true in model $1$. Therefore, we can conclude that
Eqn.~\ref{eq:family_undecidable} is unprovable under open world
assumption.  Therefore, $\ModelName{}$ is capable of handling
entailed, disproved, and unprovable axioms under open world
assumption, which empowers approximate entailment in the Family
Ontology; as more models are generated, the entailment will become
more accurate.

As the results on \textbf{Pizza} show in the first row of Table
\ref{tab:pizza_paraconsistent}, the MAE of computing axioms that
should be entailed is very close to $0$.  Also, the capability of
distinguishing entailments and unprovable axioms is `Excellent
Discrimination' according to the criterion summarized by
\cite{hosmer2013applied}, demonstrating the effectiveness of
$\ModelName$ in computing approximate entailments.

The baseline approaches \cite{bordes2013translating,yang2015embedding,dai2018novel} in Table \ref{tab:HPO} are well-established knowledge graph completion methods.
They are not able to perform semantic entailment because they are limited to triple-wise predictions without considering logical axioms.
However, as shown in the last row of Table \ref{tab:HPO}, $\ModelName$ is not only capable of computing
entailments on large real-world ontologies such as \textbf{HPO}, but
also able to distinguish between entailments and unprovable axioms
under the open world assumption. 
Note that HermiT \cite{shearer2008hermit} did not finish for this task within 24h.

\begin{table}[!t]
  	\renewcommand\tabcolsep{5pt}
  	\small
  \centering
  \caption{Paraconsistent reasoning results with $k$=1 on \textbf{Pizza}. $N_{inc}$ denotes the number of inconsistent
    axioms added.}

    \begin{tabular}{c||c||c|c|c|c}
    \multirow{2}[4]{*}{$N_{inc}$} & $\mathcal{ALC}$  & \multicolumn{4}{c}{$\ModelName{}$} \\
\cmidrule{3-6}          & Reasoners & MAE   & AUC   & AUPR  & Fmax \\
    \midrule
    \textbf{0} &    $\checkmark$    & 0.0181 & 0.8365 & 0.8804 & 0.8240 \\
    \textbf{1} &    $\times$   & 0.0186 & 0.8434 & 0.8669 & 0.8033 \\
    \textbf{5} &    $\times$   & 0.0161 & 0.8380 & 0.8849 & 0.8318 \\
    \textbf{10} &   $\times$    & 0.0158 & 0.8507 & 0.8777 & 0.8246 \\
    \midrule
    \textbf{50} &   $\times$    & 0.0914 & 0.7811 & 0.8121 & 0.7967 \\
    \textbf{100} &  $\times$     & 0.1016 & 0.7661 & 0.8075 & 0.7642 \\
    \textbf{500} &    $\times$   & 0.1516 & 0.7142 & 0.6614 & 0.7545 \\
    \textbf{1000} &   $\times$    & 0.2033 & 0.5524 & 0.5227 & 0.6667 \\
    \end{tabular}%
  \label{tab:pizza_paraconsistent}%
\end{table}%

\begin{table}[!t]
\small
  	\renewcommand\tabcolsep{2pt}
  	\small  	
  \centering
  \caption{GGI and Semantic Entailment results on HPO.}
    \begin{tabular}{c||p{0.85cm}p{0.7cm}p{0.9cm}||p{0.75cm}p{0.75cm}p{0.85cm}c}
          & \multicolumn{3}{c||}{\textbf{GGI}} & \multicolumn{4}{c}{\textbf{Semantic Entailment}} \\
\cmidrule{2-8}          & MRR   & H@3 & H@10 & MAE   & AUC   & AUPR  & Fmax \\
    \midrule
    \textbf{DistMult} & 9.6\% & 8.1\% & 14.7\% & \centering $\times$    &  \centering $\times$    &   \centering $\times$   & {\centering$\times$} \\
    \textbf{TransE} &   8.2\% & 8.3\% & 17.2\% & \centering$\times$    &   \centering $\times$   &    \centering  $\times$ & $\times$ \\
    \textbf{ConvKB} &   8.6\% & 7.9\% & 17.3\% & \centering$\times$    &   \centering $\times$   &  \centering    $\times$ & $\times$ \\
    \midrule
    $\ModelName{}$ &   \textbf{10.1\%} &   \textbf{9.7\%}    &   \textbf{20.0\%}    &    0.024   &    0.805   &   0.847    &  0.752 \\
    \end{tabular}%
  \label{tab:HPO}%
\end{table}%

\subsubsection{Paraconsistent reasoning (RQ2)}
We add explicit contradictions to the \textbf{Pizza} to test
entailment under inconsistency, i.e., paraconsistent reasoning. As
shown in Table \ref{tab:pizza_paraconsistent}, with the introduction
of a single or few inconsistent statements, symbolic \ALC{} reasoners
such as FaCT++ \cite{tsarkov2006fact++} and HermiT
\cite{shearer2008hermit} fail to compute useful entailments. However,
$\ModelName{}$ can still compute statements that would have been
entailed without the inconsistent statements added with low error, and
distinguish between entailments and unprovable axioms. Entailed axioms
have additional support within the loss function and the parts of the
models related to entailed axioms will degenerate more slowly than
other parts; this difference allows $\ModelName{}$ to compute
entailments even in the presence of inconsistency.  However,
$\ModelName$ will eventually fail to distinguish between entailed and
non-entailed statements when many contradictory statements are present
(shown in the last few rows of Table
\ref{tab:pizza_paraconsistent}). This is also a direct consequence of
the fact that $\ModelName{}$ computes approximate entailments and that
every statement is entailed by an inconsistent ontology.  We also
evaluate the improvement provided by multi-model entailment under
inconsistency (Table 3 in the Supplementary Materials).

\subsubsection{\ALC{} Enhanced prediction (RQ3)}

\begin{table}[t!]
\small
  \centering
  	\renewcommand\tabcolsep{4pt}
  \caption{PPI results on \textbf{Yeast}.}
    \begin{tabular}{c||cc||cc}
          & H@10R & H@10F & H@100R & H@100F \\
    \midrule
    ELEm  & 0.08  & 0.17  & 0.44  & 0.62 \\
    EmEL++ & 0.08  & 0.16  & 0.45  & 0.63 \\
    Onto2Vec & 0.08  & 0.15  & 0.35  & 0.48 \\
    OPA2Vec & 0.06  & 0.13  & 0.39  & 0.58 \\
    BoxEL & \textbf{0.09} & 0.20  & \textbf{0.52} & 0.73 \\
    \midrule
    $\ModelName{}$ & \textbf{0.09} & \textbf{0.23} & 0.51  & \textbf{0.75} \\
    \end{tabular}%
  \label{tab:el}%
\end{table}%

We use the exemplary \textbf{HPO} to show scalability by predicting GGIs based on phenotypic relatedness of
the genes. These can be expressed by ABox facts (triples) and thus
knowledge graph completion (KGC) methods are naturally
applicable. Representative methods include TransE
\cite{bordes2013translating}, DistMult \cite{yang2015embedding}, and
ConvKB \cite{dai2018novel}. However, background knowledge on GGIs is
expressed by \ALC{} axioms and KGC methods can only deal with triples.
To ensure fair comparisons, we only use axioms that the baseline can also express. That is, we use a subset of TBox
axioms, i.e., $C \sqsubseteq D$ (with $C, D$ limited to concept
names), along with the ABox as the training data for KGC methods.  To
further compare $\ModelName{}$ with representative ontology embedding, we predict PPIs on the {\bf Yeast} benchmark set. The models BoxEL \cite{xiong22_iswc}, EmEl++ \cite{mondal2021emel++}, and ELEm \cite{kulmanov2019embeddings} are
limited to $\mathcal{EL^{++}}$ axioms, not incorporating all axioms
provided in the \ALC{} ontology in \textbf{Yeast}. Onto2Vec \cite{smaili2018onto2vec} and OPA2Vec \cite{smaili2019opa2vec} are graph-based embedding methods not preserving logical properties.
$\ModelName{}$ uses the same ABox and the whole TBox with all \ALC{} axioms. Both GGI and
PPI predictions are formulated as approximate entailment with the
number of models $k$ set to $1$.  As the results shown in Tables
\ref{tab:HPO} and \ref{tab:el}, $\ModelName{}$ can outperform most KGC
and ontology embedding methods on most metrics for GGI and PPI
predictions. This shows 1. that the added \ALC{} axioms
are informative and helpful, and 2. that
$\ModelName$ can effectively exploit and incorporate the information
of complex \ALC{} axioms to enhance predictions. 

\section{Related work}
\subsection{Symbolic \ALC{} reasoners}
Several automated reasoners \cite{shearer2008hermit,tsarkov2006fact++}
implement sound and complete algorithms for semantic entailment over
\ALC{} ontologies.  Inconsistent ontologies have no models and
therefore all statements are semantically entailed from inconsistent
ontologies.  Consequently, they will also no longer compute
\textit{useful} entailments given inconsistencies. Several paraconsistent reasoning
approaches
\cite{schlobach2003non,flouris2008ontology,kaminski2015efficient} were
proposed to solve this issue, but discard or alter potentially crucial
knowledge \cite{troquard2018repairing}.  
Further, sound and complete symbolic approaches do
not allow for approximate semantic entailment, which may be useful
when applied to incomplete knowledge bases.  For example, sound and
complete approaches will not entail
$Father \sqsubseteq Parent \sqcap Male$ from
$Father \sqsubseteq Parent$ in the absence of
$Father \sqsubseteq Male$, even if all named instances of $Father$ are
also instances of $Male$ (in the absence of domain closure). 
$\ModelName{}$ allows for both paraconsistent and
approximate semantic entailment.

\subsection{Neural logical reasoners}
Logical query answering methods
\cite{hamilton2018embedding,ren2020query2box} reason over knowledge
graphs, which are subsets of ontologies that only include ABox axioms
of the type $R(a,b)$ and $C(a)$.  Knowledge graph completion methods
\cite{bordes2013translating,yang2015embedding,tang2022positive,hohenecker2018ontology} can be
regarded as a special case where queries are restricted to the form of
1-projection \cite{ren2020beta}. However, these methods are limited to
subsets of ontologies and thus only deal with a subproblem of semantic
entailment.  Another set of methods reason over ontologies by
generating a single model based on geometric shapes for
$\mathcal{EL^{++}}$
\cite{kulmanov2019embeddings,sun2020faithful,mondal2021emel++,peng2022description,singh2021neuro}
or \ALC{} \cite{o2021cone,leemhuis2022learning}.  These approaches are
not able to represent arbitrary models of an ontology due to
constraints of their underlying geometric shapes (see Appendix).
Hence, they are not suitable for computing approximate entailments as
they will always miss some models and therefore do not converge to
semantic entailment; they are not representation
complete. \cite{o2021cone} is capable of geometrically representing
models for ALC ontologies, but at present the applicability to
real-world scenarios, in particular involving learning including
relations, remains an open question. Logic Tensor Network (LTN)
\cite{badreddine2022logic} and its extensions
\cite{luus2021logic,wagner2022neural} enable neural first-order logic
(FOL) reasoning.  Since DLs are fragments of FOL, \ALC{} can be
handled by LTNs.
However, the direct application of LTNs to large and complex ontologies containing
many relations symbols and constants is computationally infeasible \cite{bianchi2019capabilities}.

\section{Conclusion}
We are the first to propose and implement a neural \ALC{} reasoner that approximates entailment over large and potentially inconsistent \ALC{} ontologies. 
Experimental results demonstrate that $\ModelName{}$ enables neural
networks to compute approximate entailments and can be used to
enhance knowledge base completion tasks with background knowledge.
Approximating logical models allows $\ModelName{}$ to reason over
large ontologies even in the presence of inconsistencies, both
intractable for classical \ALC{} reasoners for larger complex ontologies.

Future works will include extensions to other DLs and applications in fields like bioinformatics where many expressive ontologies exist.
Further, we can also relax the notion of maximum degree of satisfiability and use the mean degree of satisfiability based on truth degree and number of models. 
Such an approach would give rise to a different form of approximate entailment, and the interactions that can occur here are subject to future research.

\bibliographystyle{kr}
\bibliography{kr-sample}

\appendix
\clearpage





\section{Supplementary Materials}

\subsection{Reproducibility}
We provide the statistics of the datasets in Table 1 and we list the
tuned hyperparameter settings in Table \ref{tab:hyperparameter}. The Family Ontology we used is defined as Eqn. \ref{family_ontology}.
The
detailed computation steps can be found in our implementation\footnote{https://anonymous.4open.science/r/FALCON-7543} and in Algorithm 1.

\setcounter{table}{0}
\setcounter{figure}{0}
\setcounter{equation}{0}
\begin{table}[htbp]
  \centering
  \caption{Statistics of datasets.}
    \begin{tabular}{c||c|c|c|c}
    \toprule
          & \textbf{Family} & \textbf{Pizza} & \textbf{HPO} & \textbf{Yeast} \\
    \midrule
    Individual Symbols & 0     & 5     & 2504 & 5586 \\
    Concept Symbols & 10    & 99    & 30893 & 45003 \\
    Relation Symbols & 2     & 3     & 186 & 11 \\
    Abox C(a) Train & 0     & 5     & 93270 & 56934\\
    Abox R(a, b) Train & 0     & 0     & 9554 & 171740 \\
    Abox R(a, b) Test & N/A   & N/A   & 2400 & 21560\\
    Tbox Train & 25    & 676   & 64416 & 120027 \\
    Tbox Test & N/A   & 115   & 2000 & N/A \\
    \bottomrule
    \end{tabular}%
  \label{tab:stats}%
\end{table}%

\begin{table}[htbp]
  	\renewcommand\tabcolsep{2pt}
  \centering
  \caption{Tuned hyperparameter settings.}
    \begin{tabular}{c||c|c|c|c}
    \toprule
          & \textbf{Family} & \textbf{Pizza} & \textbf{HPO} & \textbf{Yeast} \\
    \midrule
    Number of models $k$ & 100    & 1-20 & 1  & 1   \\
    Contradictory axioms $N_{inc}$ & 0     & 0-1000     & N/A  & N/A   \\
    Learning rate & $1e^{-2}$   & $5e^{-3}$ & $1e^{-4}$ & $1e^{-4}$ \\
    Embedding dimension & 50        & 50    & 128  & 32  \\
    Negatives & N/A     & N/A   & 8   & 8   \\
    Batchsize R(a, b)    & N/A   & N/A   & 64   & 64  \\
    Batchsize C(a) & N/A      & N/A   & 64   & 64  \\
    Batchsize TBox & 25       & 256   & 64   & 64  \\
    Sampled individuals & 4          & 4     & 4   & 4  \\
    Created ABox axioms & 10       & 99    & 1000  & 4000 \\
    $t$-norm & Product  & Product & Product & Product \\
    \bottomrule
    \end{tabular}%
  \label{tab:hyperparameter}%
\end{table}%

\begin{equation}
\begin{split}
    &\text{Male} \sqsubseteq \text{Person},
    \text{Female} \sqsubseteq \text{Person},
    \text{Male} \sqcap \text{Female} \sqsubseteq \bot, \\
    &\text{Parent} \sqsubseteq \text{Person},
    \text{Child} \sqsubseteq \text{Person}, 
    \text{Parent} \sqcap \text{Child} \sqsubseteq \bot, \\
    &\text{Father} \sqsubseteq \text{Male},
    \text{Boy} \sqsubseteq \text{Male}, 
    \text{Father} \sqcap \text{Boy} \sqsubseteq \bot, \\
    &\text{Mother} \sqsubseteq \text{Female},
    \text{Girl} \sqsubseteq \text{Female},
    \text{Mother} \sqcap \text{Girl} \sqsubseteq \bot, \\
    &\text{Father} \sqsubseteq \text{Parent},
    \text{Mother} \sqsubseteq \text{Parent}, 
    \text{Father} \sqcap \text{Mother} \sqsubseteq \bot, \\
    &\text{Boy} \sqsubseteq \text{Child},
    \text{Girl} \sqsubseteq \text{Child}, 
    \text{Boy} \sqcap \text{Girl} \sqsubseteq \bot, \\
    &\text{Female} \sqcap \text{Parent} \sqsubseteq \text{Mother},\\
    &\text{Male} \sqcap \text{Parent} \sqsubseteq \text{Father}, \\
    &\text{Female} \sqcap \text{Child} \sqsubseteq \text{Girl},\\
    &\text{Male} \sqcap \text{Child} \sqsubseteq \text{Boy}, \\
    &\exists \text{hasChild.Person} \sqsubseteq \text{Parent}, \\
    &\exists \text{hasParent.Person} \sqsubseteq \text{Child}, \\
    & \text{Grandma} \sqsubseteq \text{Mother} 
\end{split}
\label{family_ontology}
\end{equation}
Note that we further add ABox axioms and create one named individual
for each concept name, e.g., creating the individual \textit{A child}
for concept name \textit{Child}.

\subsection{Additional Experiments}

As shown in Table \ref{tab:pizza_para_and_multi}, with the introduction of inconsistency, the improvement of \textbf{Multi} over \textbf{AVG} becomes more
significant; as contradictory statements tend to create degenerate
models (with empty concepts and relations), aggregating over multiple
models generally perform better under inconsistency since some models
still tend to preserve the memberships required for entailment.

\begin{table}[t!]
  \centering
  \caption{Reasoning performance (AUC) with 10 models and improvement
    of the multi-model semantic entailment described in Section 4.2
    across increasing inconsistency on Pizza, where \textbf{Avg}
    denotes the average results (AUC) of independent single models and
    \textbf{Multi} denotes multi-model approximate
    entailment. $N_{inc}$ denotes the number of explicitly added
    contradictions.}
    \begin{tabular}{c||c|c|c|c}
    \toprule
       $N_{inc}$   & \textbf{0} & \textbf{1}  & \textbf{10} & \textbf{50} \\
    \midrule
    \textbf{Avg}  & 0.8418 & 0.8375  & 0.8500 & 0.7884 \\
    \midrule
    \textbf{Multi} & 0.8543 & 0.8564  & 0.8706 & 0.8032 \\
    \midrule
    \textbf{Impr.($\%$)} & \textbf{3.66$\%$} & \textbf{5.04$\%$} & \textbf{5.89$\%$} & \textbf{5.13$\%$} \\
    \bottomrule
    \end{tabular}%
  \label{tab:pizza_para_and_multi}%
\end{table}%

\begin{figure}[t!]
	\centering  
	\includegraphics[width=0.45\textwidth]{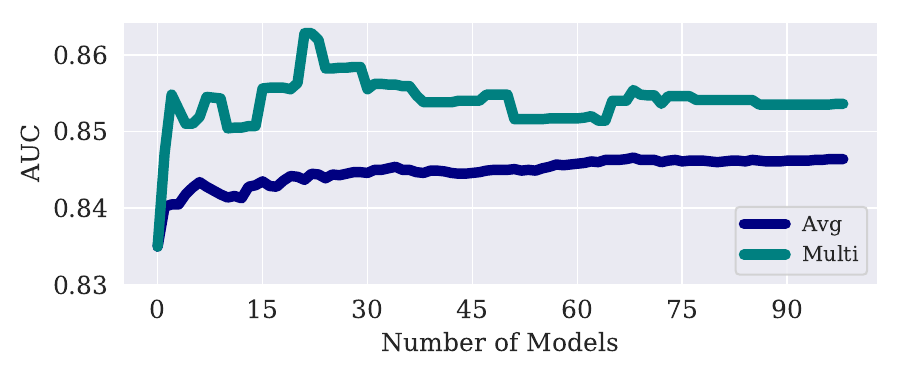}
	\caption{Experimental results of multi-model reasoning on
          Pizza. \textbf{Avg} denotes the average results (AUC) of
          independent single models. \textbf{Multi} denotes semantic
          entailment over multiple models described in Section
          \ref{multiple_model_entailment}.}
	\label{fig:pizza}
\end{figure}

We also evaluate the effectiveness of approximate entailment under
different number of models. As the results show in Figure
\ref{fig:pizza}, we observe that the approximate entailment
performance improves with a larger number of models and will
eventually converge. Although the performance will fluctuate with
fewer models, multi-model approximate entailment can consistently
improve over the single-model case, i.e., Avg in Figure
\ref{fig:pizza}.
Such results demonstrate the effectiveness of reasoning with multiple
models as well as the rationale for setting the number of models as a
hyperparameter in practice.

\subsection{Properties of $t$-norm, $t$-conorm, and fuzzy negation}
$\tnorm$ is a $t$-norm, therefore
$\tnorm:[0,1]\times[0,1]\rightarrow[0,1]$ is associative, commutative,
has $1$ as its identity element and is monotone in both arguments,
i.e., whenever $x\leq x'$ and $y\leq y'$, then
$\tnorm(x,y)\leq \tnorm(x',y')$.  We assume $\tnorm$ to be continuous,
and $\tnorm(x,y)=0$ if and only if $x=0$ or $y=0$ (and therefore
equivalent to classical semantics when $x,y\in\{0,1\}$). The
$t$-conorm is a binary operator, defined as
$\tconorm(x,y) := 1-\tnorm(1-x, 1-y)$. We assume fuzzy negation
$\fneg(x)$ to be strong, i.e., if $x<x'$ then $\nu(x)>\nu(x')$, and
involutive, i.e., $\forall x\in[0,1]:\nu(\nu(x))=x$.

In $\ModelName{}$ we consider and implement the G\"odel $t$-norm
$\tnorm(x, y)=\min \{x, y\}$, the Product $t$-norm
$\tnorm(x, y)=x \cdot y$, and the Łukasiewicz $t$-norm
$\tnorm(x, y)=\max \{x + y - 1, 0\}$; while the G\"odel and Product
$t$-norms satisfy the conditions on $t$-norms stated in Theorem 1, the
Łukasiewicz $t$-norm is not a strict $t$-norm and does not satisfy
these conditions.

\subsection{Proofs}

\setcounter{theorem}{0}

\textbf{Lemma 1} (Balls and boxes are not representation complete.)\\
    1. As the intersection of two balls is not a ball itself, the equivalence $A \sqcap B \equiv C$ may only be represented iff $A \equiv B \equiv C$ holds. This breaks given additional axioms $D\sqsubseteq B$, $D\sqcap C \sqsubseteq \bot$, $D$ non-empty.\\
    2. (Axis-aligned) boxes \cite{peng2022description} cannot represent the axiom $X \equiv A \sqcap B$, $Y \equiv B \sqcap C$, $Z \equiv A\sqcap C$; $X \sqcap Y \sqsubseteq \bot$, $Y \sqcap Z \sqsubseteq \bot$, $X \sqcap Z \sqsubseteq \bot$.\\
\begin{theorem}[Faithfulness]
  Let $\TBox$ be a TBox and $\ABox$ an ABox in \ALC{} over signature
  $\Sigma = (\mathbf{C}, \mathbf{R}, \mathbf{I} = \mathbf{I_n} \cup
  \mathbf{I_{\Re^n}})$. If $\mathcal{L}=0$ and the $t$-norm $\tnorm$
  satisfies $\tnorm(x,y)=0$ if and only if $x=0$ or $y=0$, then the
  interpretation $\I$ with $C^\I = f_\I(f_e(C))$ for all
  $C \in \mathbf{C}$, $R^\I = f_\I(f_e(C))$ for all
  $R \in \mathbf(R)$, and $a^\I = a$ for all $a \in \mathbf{I}$ is a
  classical model of $\TBox$ and $\ABox$ with domain $\Delta = \Re^n$.
\end{theorem}
\begin{proof}
  By definition, $f_{mod} = f_e \circ f_\I$ generates an interpretation. We need
  to show that such an interpretation is a model, i.e., all axioms in
  $\TBox \cup \ABox$ are true in the interpretation $\I$ generated by
  $f_e \circ f_\I$ if $\mathcal{L}$ is $0$.  For concept descriptions
  $C,D$, we need to show that, if the loss is $0$, in the
  interpretation $\I$, the degree of membership for all individuals in
  $(C \sqcap \neg D)^\I$ is $0$ for every $C \sqsubseteq D \in \TBox$.
  $\mathcal{L} = 0$ implies that $\mathcal{L}_{\mathcal{T}} = 0$.
  Because membership is always positive, i.e.,
  $m_\cdot(\cdot)\geq 0$, $m_{(C\sqcap\neg D)}(x)=0$ is true for
  all $C\sqsubseteq D \in \TBox$ and $x \in E$.

  Consider $m_{(C\sqcap\neg D)}(x)=0$ for some $x\in E$.
  Let $C$ and $D$ be concept names (induction start), then, by the
  condition that $\tnorm(x_1,x_2) = 0$ iff $x_1=0$ or $x_2=0$, either
  $m_C(x) = 0$ or $m_{\neg D}(x) = 0$ for any $x$; by
  monotonicity of fuzzy negation and Eqn. \ref{eqn:negation}, either
  $m_C(x) = 0$ or $m_D(x) = 1$.  Let $C$ and $D$ be concept
  descriptions (induction step).  If $C = \neg A$ then we minimize
  $\fneg (m_A(x))$; by monotonicity of $\fneg$ this maximizes
  $m_A(x)$, i.e., $m_A(x) = 1$ for all $x$; analogously for $D$.
  If $C = A \sqcap B$ then we minimize
  $\tnorm(m_A(x), m_B(x))$. $\tnorm$ is commutative and
  monotone, and minimizing the $t$-norm will minimize at least one
  conjunct; the value is minimal ($0$) only when one conjunct is $0$
  (and maximal when both are $1$ for maximizing $D^\I$).  Let
  $C = A \sqcup B$; the $t$-conorm $\tconorm$ is minimal when
  membership in both $A^\I$ and $B^\I$ are $0$ (either $m_A(x)=1$
  or $m_B(x)=1$ when maximizing $D^\I$).  For $C = \exists R.A$, we
  rely on Eqn. \ref{eqn:existential} for generating the interpretation
  of $R$; Eqn. \ref{eqn:existential} relies on an MLP and combines
  with $A^\I$ through a $t$-norm; $C^\I$ is minimized if $m_A(x)=0$
  (induction hypothesis), or the $\max_{y \in \Delta}$ (which runs
  over finite subsets of $\Delta$ sampled from $\mathbf{I}$) is $0$
  when no individual $y$ stands in relation $R$ to some member of
  $A^\I$; because we use $\tnorm$ in Eqn. \ref{eqn:existential}, this
  entails $m_R((x,y)) = 0$.  The argument runs analogously for
  $C = \forall R.A$, and similarly for maximizing $D^\I$.
  
  If $\mathcal{L} = 0$ then $a^\I \in C^\I$ for all $C(a) \in
  \ABox$. $\mathcal{L} = 0$ implies
  $ \mathcal{L}_{\mathcal{A}_1} = \frac{1}{|\mathcal{A}_1|} \sum_{C(e)
    \in \mathcal{A}} (1-(m(e, C^\I))) = 0$. If $C$ is a concept name,
  the degree of membership is $1$ for all $a^\I$ with
  $C(a) \in \ABox$. If $C$ is a concept description, the statement
  follows inductively as for TBox axioms and $\mathcal{L_\TBox}$.

  If $\mathcal{L} = 0$ then $(a^\I,b^\I) \in R^\I$ for all
  $R(a,b) \in \ABox$. In the absence of axioms for relations, this
  follows directly from the definition of
  $ \mathcal{L}_{\mathcal{A}_2}$.
\end{proof}

\begin{theorem}[Representation completeness]
  Let $\TBox$ be a TBox and $\ABox$ an ABox in \ALC{} over signature
  $\Sigma = (\mathbf{C}, \mathbf{R}, \mathbf{I} = \mathbf{I_n} \cup
  \mathbf{I_{\Re^n}})$. If $\I$ is a finite model of
  $\TBox \cup \ABox$ then there exists an $f_e$ and $f_\I$ such that
  $\mathcal{L}\leq \epsilon$ for any $\epsilon>0$.
\end{theorem}
\begin{proof}
  Given a model $\I$, we need to show that we can find $f_e$ and
  $f_\I$ so that $\mathcal{L} \leq \epsilon$.  $f_e$ is an embedding
  function and $f_\I$ relies on MLPs as universal approximators
  \cite{hornik1989multilayer}, and therefore it is possible to make
  any assignment to degrees of membership for any concept and relation
  name (Eqns. \ref{eqn:fsconcept} and \ref{eqn:fsrelation}). Let $U$
  be the universe of $\I$.  We select $f_\I$ to assign the following
  degrees of membership to embeddings $x \in \Re^n$ (either generated
  through $f_e$ or through sampling from $\mathbf{I_{\Re^n}}$):
  \begin{itemize}
  \item if $x \in C^\I$ then $m_C(x) = 1 - \epsilon'$, otherwise
    $m_C(x) = \epsilon'$,
  \item if $(x,y) \in R^\I$ then $m_C((x,y)) = 1 - \epsilon'$,
    otherwise $m_R((x,y)) = \epsilon'$.
  \end{itemize}
  By this construction, $\mathcal{L}_{\mathcal{A}_2}$ will at most be
  $\epsilon'$ and $\mathcal{L}_{\mathcal{A}_1}$ at most $\epsilon'$
  for all axioms of the type $C(a)$ where $C$ is a concept name;
  similarly, $\mathcal{L_\TBox}$ will be at most $\epsilon'$ for
  axioms $C \sqsubseteq D$ where $C$ and $D$ are concept names. If $C$
  and $D$ are concept descriptions, the losses are also bounded by
  $\epsilon'$ due to the definition of degrees of membership for
  concept descriptions
  (Eqns. \ref{eqn:intersection}--\ref{eqn:universal}). Consequently,
  $\mathcal{L} \leq \epsilon \leq 3 \cdot \epsilon'$.
\end{proof}

\end{document}